# Automated Detection and Analysis of Power Words in Persuasive Text Using Natural Language Processing

**Mr. Sahil Garje**

School of Computing Science and Engineering, VIT Bhopal University, Madhya Pradesh, India
Email: sahilgarje@gmail.com

## Abstract

Power words are terms that evoke strong emotional responses and significantly influence readers' behavior, playing a crucial role in fields like marketing, politics, and motivational writing. This study proposes a methodology for the automated detection and analysis of power words in persuasive text using a custom lexicon created from a comprehensive dataset scraped from online sources. A specialized Python package, *The Text Monger*, is created and employed to identify the presence and frequency of power words within a given text. By analyzing diverse datasets, including fictional excerpts, speeches, and marketing materials, the aim is to classify and assess the impact of power words on sentiment and reader engagement. The findings provide valuable insights into the effectiveness of power words across various domains, offering practical applications for content creators, advertisers, and policymakers looking to enhance their messaging and engagement strategies.

## Keywords

Text Analysis, Power Words, Sentiment Analysis, Text Categorization, Natural Language Processing

# 1. Introduction

Power words are compelling terms strategically chosen to evoke emotional responses and drive action. These words play a crucial role in various forms of communication, including marketing, persuasive writing, and public speaking. For instance, a study in *Psychology Today*[1] highlights that power words can shape the filters through which individuals perceive reality, ultimately influencing their behaviors. Bamberg (1997)[2] emphasizes the complex interplay between language and emotions, illustrating how language can both reflect and shape feelings. Despite their significance, existing methodologies for detecting power words remain limited, often relying on manual identification rather than automated approaches.

This paper aims to fill this gap by developing a systematic, lexicon-based tool to identify and categorize power words in textual data. A comprehensive dataset of power words was compiled by scraping online sources, allowing for the creation of a specialized lexicon. Custom functions were developed to analyze text for the presence and distribution of these power words. *The Text Monger*, a Python package created for this purpose, also incorporates various text analysis tools, including readability scoring, named entity recognition (NER), and sentiment analysis. Through extensive data collection from sources like Project Gutenberg for fictional excerpts, American Rhetoric for speeches, and marketing materials from major tech companies, this research explores how specific word choices shape emotional responses and sentiment across different contexts. The results reveal significant trends in readability levels, power word distributions, and sentiment metrics, providing valuable insights into the effectiveness of language strategies in communication. By addressing questions regarding how power words influence sentiment across genres, this research aspires to enhance the understanding of linguistic strategies, offering actionable insights for marketers, writers, and communicators seeking to improve their messaging and engage more effectively with audiences.

# 2. Literature Review

The significance of language in shaping perceptions and decision-making processes has been well-documented across various fields. Nickel et al. (2017) conducted a systematic review examining how different terminology for the same medical condition influences management preferences**[3]**. Their findings suggest that terminology can significantly affect patients' treatment choices and emotional responses. When more medicalized or precise terms were employed, there was a tendency towards preferences for more invasive management options, as well as increased anxiety and perceived severity of the condition. This underscores the critical role of language in influencing emotional and psychological outcomes, which is foundational to understanding the power of words in communication contexts.

Building on the idea of language shaping perceptions, the psychological foundations of power words draw from cognitive psychology, where specific language choices are linked to emotional resonance and behavior. Power words are strategically selected to evoke strong emotional responses, enhancing the persuasive impact of messages (Bamberg, 1997). They serve as linguistic tools that can not only reflect but also shape emotions, reinforcing their effectiveness in domains like marketing and public speaking. Silberstein (2002) highlights how language has been wielded as a powerful instrument in political discourse, particularly following the events of September 11th**[4]**. In her analysis, she discusses how narratives constructed around terrorism influenced public sentiment and polarized opinions during a crisis, reflecting language's role in constructing reality and consensus.

Taillard (2000) further explores the integration of pragmatics and social psychology in understanding persuasive communication, particularly within marketing contexts**[5]**. She asserts that successful persuasion encompasses two primary goals: ensuring comprehension and fostering belief in the audience. By combining insights from both fields, Taillard proposes a model that delineates the process of persuasion from the speaker's intention to potential attitude changes in the hearer. This integrated approach enriches the understanding of how persuasive language operates in optimized contexts like marketing, emphasizing the need for interdisciplinary collaboration in studying communication dynamics.

Furthermore, research indicates that power words enhance engagement and motivation in communication. Words that evoke strong imagery or emotional reactions can lead to increased persuasion and action (Cialdini, 2009)**[6]**. This aligns with Nickel et al.'s findings, which demonstrate that language not only conveys information but also frames the way individuals perceive that information, influencing their preferences and decisions.

Automated detection of power words within textual data presents a novel area of exploration in Natural Language Processing (NLP). By developing a systematic approach to identify and categorize these words, this study aims to provide valuable insights into how specific linguistic choices impact emotional responses and overall sentiment in various genres of text. Understanding the contextual application of power words can lead to practical implications for content creators, advertisers, and policymakers, facilitating more effective messaging strategies. This literature review highlights the need for further exploration into automated detection of power words, paving the way for a deeper understanding of their impact across diverse textual contexts.

# 3. Methodology

### 3.1 Data Collection

The dataset utilized for this research was compiled from three distinct sources: fictional excerpts from *Project Gutenberg* **[7][8][9][10][11]**, speeches from *American Rhetoric* **[12][13][14][15][16]**, and marketing excerpts from major tech companies, including *Amazon* and *Apple* **[17][18][19][20][21]**. The fictional excerpts were selected to provide diverse narrative styles, while the speeches were chosen for their rhetorical significance and impact on audiences. The marketing excerpts were included to analyze contemporary persuasive techniques employed by leading corporations. Each source contributed a rich variety of texts, enabling a comprehensive analysis across different contexts. The figures **3.3**, **3.4(b)**, **3.5**, and **3.6** show a sample run of the program on *Martin Luther Jr.'s 'I Have a Dream'* speech acquired through *American Rhetoric*.

### 3.2 Text Processing

To analyze the collected texts, a Python package titled *[The Text Monger](#)* was developed. This package includes various text analysis tools, such as readability scoring, power words distribution, named entity recognition (NER), and sentiment analysis. The text processing involved cleaning the data by removing punctuation, converting it to lowercase, and tokenizing the words for easier analysis.

### 3.3 Readability Analysis

Readability metrics were calculated using the features integrated within *The Text Monger*. This involved assessing the texts against various indices, including the Flesch Reading Ease, Gunning Fog Index, SMOG Index, Coleman-Liau Index, and Dale-Chall Readability Score, utilizing the capabilities of the [textstat](#) library. These metrics provided insights into the complexity of the texts and their accessibility to a broad audience.

**Fig. 3.3** Readability Analysis

| Metric | Score |
|---|---|
| Reading ease | Standard |
| Reading level | Grade 8.8 |
| Smog index | Grade 11.0 |
| Gunning Fog index | Grade 10.52 |
| Coleman-Liau index | Grade 8.01 |
| Automated Readability index | Grade 9.8 |
| Dale-Chall Readability score | 7.27 |
| Text standard | 10th and 11th grade |

**3.4 Power Words Distribution**

The analysis of power words was conducted using a pre-compiled dataset of power words categorized into types such as Greed, Encouragement, Safety, Anger, Lust, Fear, and Forbidden. These power words were sourced from two websites: OptinMonster [22] and SmartBlogger [23]. A Python function was developed to scan through the texts and identify occurrences of these power words. The function counted how many times words from each category appeared in the text, enabling a detailed understanding of the emotional and psychological triggers employed in marketing and speeches.

**Figure 3.4(a):** Bar graph representing the structure of acquired data

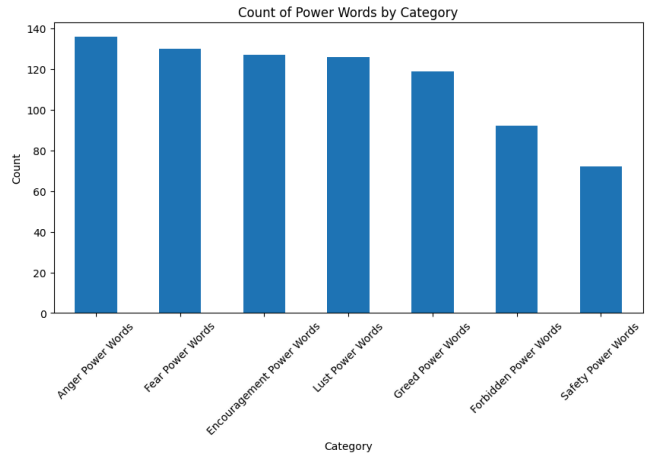

**Figure 3.4(b):** Power Words Distribution

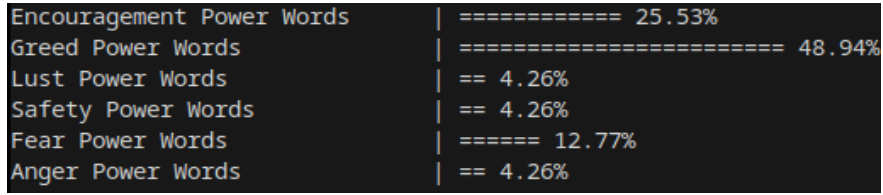

**3.5 Sentiment Analysis**

Sentiment analysis was performed using the TextBlob library incorporated within *The Text Monger*. Each text was analyzed for its sentiment polarity, which ranges from -1 (negative) to +1 (positive), with 0 indicating neutrality. The analysis included a correlation between the presence of power words and the overall sentiment polarity of the texts, allowing for insights into how specific categories of power words contribute to the emotional tone of marketing materials, speeches, and fictional narratives.

**Figure 3.5:** Sentiment Analysis

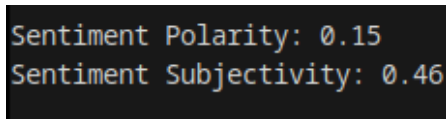

## 3.6 Named Entity Recognition (NER)

Named Entity Recognition was utilized, incorporating the spaCy library, to extract and highlight significant entities (like product names, organizations, and notable figures) from the texts. This analysis reinforced the brand identity in marketing materials and the credibility of speakers in speeches, showcasing their relevance and authority in respective contexts.

**Figure 3.6:** NER Recognition

I am happy to join with you [today DATE] in what will go down in history as the greatest demonstration for freedom in the history of our nation.

[Five CARDINAL] [score years ago DATE], a great [American NORP], in whose symbolic shadow we stand [today DATE], signed [the Emancipation Proclamation LAW]. This momentous decree came as a great beacon light of hope to [millions CARDINAL] of Negro slaves who had been seared in the flames of withering injustice. It came as a joyous daybreak to end [the long night TIME] of their captivity.

But [one hundred years later DATE], the [Negro ORG] still is not free. [One hundred years later DATE], the life of the [Negro ORG] is still sadly crippled by the manacles of segregation and the chains of discrimination. [One hundred years later DATE], the [Negro ORG] lives on a lonely island of poverty in the midst of a vast ocean of material prosperity. [One hundred years later DATE], the [Negro ORG] is still languished in the corners of [American NORP] society and finds himself an exile in his own land. And so we've come here [today DATE] to dramatize a shameful condition.

In a sense we've come to our nation's capital to cash a check. When the architects of our republic wrote the magnificent words of the [Constitution LAW] and the Declaration of Independence, they were signing a promissory note to which every [American NORP] was to fall heir. This note was a promise that all men, yes, black men as well as white men, would be guaranteed the "unalienable Rights" of "[Life, Liberty WORK_OF_ART] and the pursuit of [Happiness ORG]." It is obvious [today DATE] that [America GPE] has defaulted on this promissory note, insofar as her citizens of color are concerned. Instead of honoring this sacred obligation, [America GPE] has given the [Negro ORG] people a bad check, a check which has come back marked "insufficient funds."

But we refuse to believe that the bank of justice is bankrupt. We refuse to believe that there are insufficient funds in the great vaults of opportunity of this nation. And so, we've come to cash this check, a check that will give us upon demand the riches of freedom and the security of justice.

We have also come to this hallowed spot to remind [America GPE] of the fierce urgency of Now. This is no time to engage in the luxury of cooling off or to take the tranquilizing drug of gradualism. Now is the time to make real the promises of democracy. Now is the time to rise from the dark and desolate valley of segregation to the sunlit path of racial justice. Now is the time to lift our nation from the quicksands of racial injustice to the solid rock of brotherhood. Now is the time to make justice a reality for all of God's children.

# 4. Results

## 4.1 Marketing analysis

The readability analysis of the collected texts indicates a generally accessible standard across various marketing materials. The average reading level is approximately **Grade 9.4**, reflecting a moderate complexity that caters to a broad audience, including high school students. The metrics suggest that while the texts are mostly understandable, they contain enough complexity to engage more informed readers.

**Key Readability Metrics**:

- **Average Reading Ease**: Standard
- **Average Reading Level**: Grade **9.4**
- **Smog Index**: Grade **11.3**
- **Gunning Fog Index**: Grade **10.8**
- **Coleman-Liau Index**: Grade **10.3**
- **Dale-Chall Readability Score**: **7.52**

**Power Words Distribution**

The analysis reveals a strong emphasis on **greed power words**, which average around **83.33%** across the texts. This indicates a marketing strategy heavily focused on consumer benefits and incentives. The presence of other power words, such as encouragement, safety, and forbidden words, shows varying degrees of engagement with emotional and psychological triggers.

**Average Power Words Distribution**:

- **Greed Power Words**: **83.33%**
- **Encouragement Power Words**: **9.73%**
- **Safety Power Words**: **5.33%**
- **Forbidden Power Words**: **2.67%**

**Sentiment Analysis**

Sentiment analysis across the texts reflects a slightly positive tone, with an average sentiment polarity of **0.23**. This suggests that the marketing materials are designed to promote a favorable perception of the products or services. The average subjectivity score of **0.45** indicates a balanced mix of objective information and subjective language, promoting brand credibility while maintaining consumer engagement.

**Average Sentiment Metrics**:

- **Sentiment Polarity**: **0.23**
- **Sentiment Subjectivity**: **0.45**

**Named Entity Recognition (NER)**

The use of specific product names in marketing texts reinforces brand identity and encourages consumer recognition. Each text prominently features product launches and updates, showcasing the companies' innovation and relevance in a competitive market.

## 4.2 Speech Analysis

The readability analysis of the collected speeches indicates a generally accessible standard with some variation in complexity. The average reading level is approximately **Grade 9.0**, reflecting a moderate complexity that caters to a wide audience, including high school students and adults. The metrics suggest that while the speeches are mostly understandable, they contain enough complexity to engage more informed listeners.

**Key Readability Metrics:**

- **Average Reading Ease:** Standard
- **Average Reading Level:** Grade 9.0
- **Smog Index:** Grade 11.0
- **Gunning Fog Index:** Grade 10.7
- **Coleman-Liau Index:** Grade 8.4
- **Dale-Chall Readability Score:** 7.3

**Power Words Distribution**

The analysis reveals a significant emphasis on greed power words, which average around 52.4% across the speeches. This indicates a rhetorical strategy heavily focused on ambition and desire. The presence of other power words, such as encouragement, safety, and fear words, shows varying degrees of engagement with emotional and psychological triggers.

**Average Power Words Distribution:**

- **Greed Power Words:** 52.4%
- **Encouragement Power Words:** 18.7%
- **Safety Power Words:** 9.5%
- **Anger Power Words:** 6.7%
- **Lust Power Words:** 4.3%
- **Fear Power Words:** 5.8%
- **Forbidden Power Words:** 5.5%

**Sentiment Analysis**

Sentiment analysis across the speeches reflects a slightly positive tone, with an average sentiment polarity of 0.13. This suggests that the speeches are designed to promote a favorable perception of the speakers' messages. The average subjectivity score of 0.44 indicates a balanced mix of objective information and subjective language, which supports engagement while maintaining credibility.

**Average Sentiment Metrics:**

- **Sentiment Polarity:** 0.13
- **Sentiment Subjectivity:** 0.44

**Named Entity Recognition (NER)**

The use of specific references to historical contexts and notable figures in the speeches reinforces the speakers' credibility and encourages audience recognition. Each speech prominently features themes of ambition, social justice, and resilience, showcasing the speakers' relevance and authority in their respective contexts.

## 4.3 Fictional Excerpts Analysis

The readability analysis of the selected fictional excerpts indicates a varied complexity across the texts. The average reading level is approximately **Grade 7.5**, reflecting a range suitable for a middle to high school audience.

**Key Readability Metrics:**

- **Average Reading Ease:** Fairly easy
- **Average Reading Level:** Grade 7.5
- **Smog Index:** Grade 9.1
- **Gunning Fog Index:** Grade 9.8
- **Coleman-Liau Index:** Grade 8.22
- **Dale-Chall Readability Score:** 7.15
- **Text Standard:** 7th to 9th grade

**Power Words Distribution** The analysis reveals a diverse use of power words, with an average emphasis on greed power words, which averages around **57.53%** across the texts. This indicates a narrative strategy that often appeals to desire, supplemented by various emotional triggers.

**Average Power Words Distribution:**

- **Greed Power Words:** 57.53%
- **Lust Power Words:** 12.77%
- **Fear Power Words:** 12.23%
- **Encouragement Power Words:** 11.06%
- **Anger Power Words:** 8.00%
- **Forbidden Power Words:** 2.02%
- **Safety Power Words:** 2.25%

**Sentiment Analysis** Sentiment analysis across the excerpts reflects a slightly positive tone, with an average sentiment polarity of **0.05**. This suggests that the narratives are designed to promote a generally favorable perception of characters and situations, though they also present complex emotional themes. The average subjectivity score of **0.50** indicates a balanced mix of objective storytelling and subjective interpretation.

**Average Sentiment Metrics:**

- **Sentiment Polarity:** 0.05
- **Sentiment Subjectivity:** 0.50

**Named Entity Recognition (NER)** The use of specific character names and settings is prevalent across the texts, reinforcing narrative identities and enhancing reader engagement with the plots.

# 5. Conclusions

This study successfully demonstrates the effectiveness of automated detection and analysis of power words across various forms of text, including marketing materials, speeches, and fictional excerpts. The findings reveal a consistent emphasis on greed power words, which significantly influence sentiment and reader engagement. Marketing texts exhibited a high average reading level, with greed power words accounting for 83.33% of the language used, indicating a strategic focus on consumer benefits. Similarly, speeches displayed a notable presence of greed and encouraging power words, contributing to a slightly positive sentiment tone, while fictional excerpts showcased diverse emotional triggers through their power word distribution.

The readability analyses highlighted that the texts are generally accessible, catering to a wide audience, while the sentiment analysis indicated a balanced mix of objective and subjective language. By utilizing a custom lexicon for power word detection, this research contributes valuable insights into the impact of word choice on emotional responses and engagement.

Overall, the findings underscore the importance of power words as a strategic tool in communication, providing practical applications for marketers, writers, and public speakers.

While this approach demonstrates strengths, it also has limitations, including potential biases in the selected power words and challenges in sentiment analysis that may not fully capture the complexity of human emotion. Future research could build upon this foundation by exploring the automation of power word detection in diverse contexts such as social media, political speeches, or advertising campaigns, and by integrating additional linguistic features to enhance analysis. Addressing these limitations will provide a more nuanced understanding of power words and their impact in communication.